\documentclass[letterpaper]{article} 
\usepackage{aaai2026}  
\usepackage{times}  
\usepackage{helvet}  
\usepackage{courier}  
\usepackage[hyphens]{url}  
\usepackage{graphicx} 
\usepackage{natbib}  
\usepackage{caption} 
\title{SCALE-LoRA: Auditing Post-Retrieval LoRA Composition with Residual Merging and View Reliability}

\author{
    Shuaipeng Zhou\textsuperscript{\rm 1}\textsuperscript{\rm 2},
    Yu Zhang\textsuperscript{\rm 1}\textsuperscript{\rm 2}\thanks{Corresponding author.}
}
\affiliations{
    \textsuperscript{\rm 1}School of Computer Science and Technology, Shenyang University of Chemical Technology\\
    \textsuperscript{\rm 2}Liaoning Key Laboratory of Intelligent Technology for Chemical Process Industry\\
    z2025525@stu.syuct.edu.cn, zytemb@163.com 
}

\begin{document}

\maketitle

\begin{abstract}
Libraries of Low-Rank Adaptation (LoRA) adapters are becoming a practical by-product of parameter-efficient adaptation. Once such adapters accumulate, a natural question is no longer how to train one adapter for one task, but how to reuse an open pool of adapters for a new task given only a small support set. Prior work has shown that LoRA modules can be composed at the task level and dynamically selected at the instance level. However, open-pool LoRA reuse is not automatic: retrieving relevant adapters does not guarantee that their parameter updates are compatible, and composing adapters does not guarantee reliable outputs.

We introduce the Sparse-Composition Agreement Layer (SCALE), a post-retrieval audit and composition framework for open-pool LoRA reuse. SCALE contains a deployable 1.0$\times$ merge path, Layer-Adaptive Sparse Residual Composition (LASRC), and a higher-cost reliability-analysis layer for multi-view disagreement. LASRC addresses merge interference by preserving a linear anchor while residualizing block-wise adapter update directions. The reliability layer treats disagreement among sparse composition views as an observable uncertainty signal and compares agreement, support-loss proxy selection, and oracle headroom under explicit path cost. In matched FLAN-T5-Large, BIG-Bench Hard (BBH), and 97-LoRA experiments, LASRC gives a directional single-view gain under fixed retrieval, while SCALE-support is reported as a query-label-free 3.0$\times$ reliability-analysis variant rather than as a calibrated or throughput-equivalent selector. Protocol-distinct BBH-8 validation shows the same qualitative trend on three decoder-only backbones. Detailed scores, paired audits, and path-cost records are reported in the experimental section.
\end{abstract}

\begin{figure*}[!t]
\centering
\includegraphics[width=\textwidth]{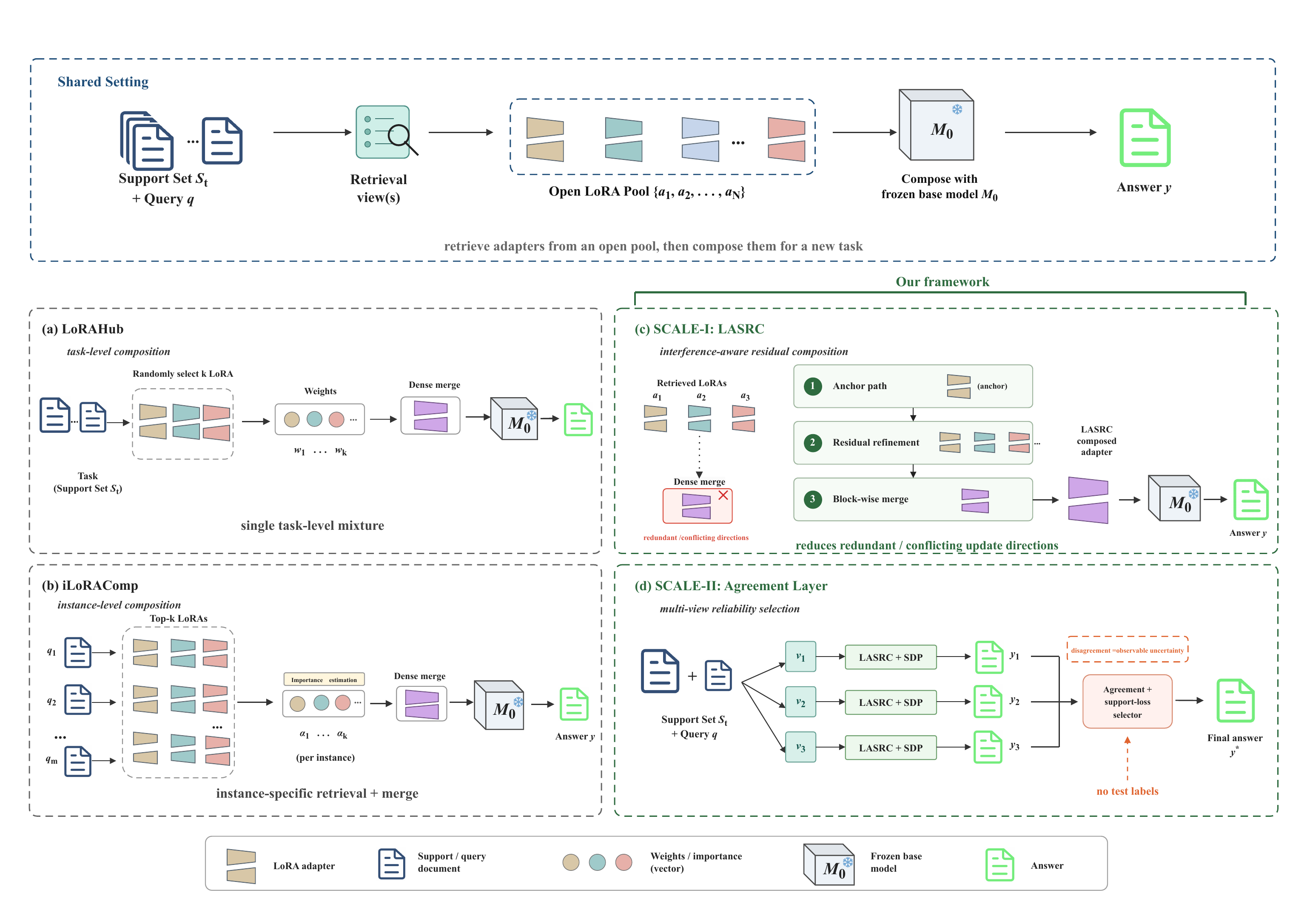}
\caption{Post-retrieval reliability in open-pool LoRA reuse. Prior methods mainly decide which adapters to retrieve and how to weight them before dense composition. SCALE separates the remaining post-retrieval problem into two bottlenecks: merge interference, where relevant adapters may not be parameter-compatible, and view reliability, where several plausible sparse composition views may disagree. LASRC addresses the first bottleneck with anchor-preserving residual composition, while SCALE analyzes the second through agreement, support-loss proxy selection, and oracle headroom under explicit path cost.}
\label{fig:open-pool-lora}
\end{figure*}

\section{Introduction}

Large language models have become general-purpose infrastructure for reasoning, coding, knowledge-intensive generation, and human-AI interaction. However, adapting such models to continuously emerging tasks remains expensive. In realistic deployment, new target tasks often differ from the source tasks on which existing adapters were trained, while collecting sufficient data or training a new adapter for every target task is impractical. This motivates cross-task LoRA reuse: instead of training a new adapter from scratch, we aim to reuse and compose adapters trained on diverse source tasks to support unseen or low-resource target tasks.

Full fine-tuning is increasingly unattractive in this setting because it updates all model parameters and creates a full model copy for every target task. Low-Rank Adaptation (LoRA) offers a different regime: it freezes the pretrained weights and injects trainable low-rank matrices into Transformer modules \cite{hu2022lora}. The original LoRA study reports up to a 10,000-fold reduction in trainable parameters and roughly a 3-fold reduction in graphics processing unit (GPU) memory requirements on GPT-3 175B while retaining competitive downstream performance. Just as importantly for cross-task reuse, LoRA packages task adaptation into compact modules that can be stored, shared, retrieved, and composed without modifying the backbone.

Once many LoRA adapters have been trained, the research question shifts from single-task adaptation to adapter-pool reuse. LoRAHub is an early representative of this shift: it studies cross-task generalization by dynamically composing multiple LoRA modules for an unseen task with only a few examples and without adding new model parameters or gradient updates \cite{huang2024lorahub}. This establishes LoRA adapters as composable capability units rather than isolated task artifacts. AdapterSoup similarly demonstrates that averaging modular adapters can improve generalization in compatible settings \cite{chronopoulou2023adaptersoup}.

Subsequent work has pushed composition from task-level reuse toward instance-level reuse. iLoRAComp argues that a single task-level mixture can miss input-specific requirements, and therefore selects and weights multiple task LoRAs for each input instance; it reports results above LoRAHub on 16 of 27 public tasks \cite{wang2024instance}. LoGo continues this line by studying instance-level dynamic LoRA selection and merging, including top-1 selection, uniform merging, and signal-weighted merging variants \cite{lee2025logo}. These studies collectively show that LoRA reuse is valuable and that the relevant unit of decision can be smaller than a whole task.

Existing open-pool LoRA reuse pipelines mostly optimize the pre-merge decision: which adapters should be retrieved and how their coefficients should be assigned. We argue that this leaves a post-retrieval reliability gap. After candidate adapters have been retrieved, the system must still determine whether their updates are merge-compatible and how to compare plausible composed views when they disagree.

Figure~\ref{fig:open-pool-lora} summarizes this positioning. LoRAHub performs one task-level mixture, and iLoRAComp moves retrieval and weighting to the instance level, but both ultimately rely on dense merging once candidate adapters have been selected. SCALE studies two post-retrieval bottlenecks: merge interference, where relevant adapters may occupy overlapping or conflicting update directions, and view disagreement, where several plausible composition views produce different answers.

SCALE therefore separates merge compatibility from view reliability. Layer-Adaptive Sparse Residual Composition (LASRC) addresses the first bottleneck as a single-path residual merge operator. The multi-view layer addresses the second by comparing answer agreement, support-loss proxies, and oracle headroom under explicit path cost. We report these components with separate claim boundaries rather than treating multi-view aggregation as a throughput-equivalent replacement for single-view composition.

Our central response is a post-retrieval audit centered on one deployable operator and one higher-cost diagnostic layer. LASRC keeps the iLoRAComp-style linear merge as an anchor, but residualizes support-weighted LoRA directions within each transformer block before forming the composed update. It does not claim that parameter-space orthogonality implies semantic independence; instead, it uses residualization as an operational mechanism to reduce repeated occupancy of similar update subspaces.

SCALE further uses Stochastic Delta Pruning (SDP) as a seeded sparse-view construction mechanism inside the multi-view layer. SDP perturbs adapter deltas before composition and is analyzed as a sparse-view control rather than as a standalone pruning method. The resulting views are then compared by agreement, support-loss weighting, and oracle headroom.

We evaluate LASRC and SCALE in a matched FLAN-T5-Large, BIG-Bench Hard (BBH), and 97-LoRA setting, and further validate the method on decoder-only backbones under a LoGo-style BBH-8 protocol. The matched experiments use the same backbone, adapter pool, support size, retrieval family, decoding path, answer normalization, and evaluation script across compared methods. Under this protocol, LASRC gives a small positive fixed-retrieval single-view merge effect relative to the iLoRAComp-style linear merge. SCALE-support is analyzed as a higher-cost reliability variant rather than as a throughput-equivalent replacement, and decoder-only BBH-8 validation is reported separately as protocol-distinct evidence.

We make five contributions:

\begin{itemize}
\item We propose LASRC, a fixed-retrieval residual merge operator that preserves a linear anchor while separating block-wise residual update directions to reduce redundant reuse of similar LoRA subspaces.
\item We introduce SDP as a seeded sparse-view construction mechanism that perturbs adapter deltas before composition, exposing when sparse control helps or hurts reuse.
\item We develop SCALE as a reliability-analysis layer that compares sparse composition views through agreement, support-loss weighting, and oracle headroom under explicit path cost.
\item We combine LASRC, SDP, and SCALE into one cross-task LoRA reuse pipeline for adapting to unseen target tasks without training a new adapter.
\item We provide matched FLAN-T5-Large/BIG-Bench Hard/97-LoRA experiments, default model-merging sanity checks, explicit path-cost reporting, and protocol-distinct decoder-only validation.
\end{itemize}

\section{Problem Formulation}

\paragraph{Source LoRA pool.}
We follow the cross-task adapter-reuse setting of LoRAHub and iLoRAComp. Let $\mathcal{S}=\{T_1,\ldots,T_N\}$ be source tasks. For each $T_i$, a Low-Rank Adaptation adapter has already been trained on a frozen backbone model $M_0$ with parameters $\phi$. The resulting adapter pool is
\begin{equation}
\mathcal{A}=\{a_i\}_{i=1}^{N},\qquad
a_i=\{\theta_{i,k}\}_{k\in\mathcal{K}},
\label{eq:adapter-pool}
\end{equation}
where $\mathcal{K}$ indexes the LoRA modules shared by all adapters and $\theta_{i,k}$ is the effective LoRA update tensor of adapter $a_i$ at module $k$. During reuse, $\phi$ is frozen and no new target-task adapter is trained.

\paragraph{Unseen target task.}
At adaptation time, the system receives an unseen target task $T$ with a small labeled support set and an unlabeled query set:
\begin{equation}
\mathcal{D}^{sup}_{T}=\{(x_j,y_j)\}_{j=1}^{K},\qquad
\mathcal{D}^{qry}_{T}=\{x_j\}_{j=1}^{M}.
\label{eq:support-query}
\end{equation}
The support set may be used for retrieval, adapter weighting, sparse-view construction, and reliability estimation. Query labels are used only after predictions are fixed, except in explicitly marked oracle diagnostics. The task is to build a target-conditioned adapter state from $\mathcal{A}$ for task $T$ without updating $\phi$.

\paragraph{Retrieval and composition.}
A retrieval rule selects a small candidate subset from the pool. In the multi-view case, each view $v\in\mathcal{V}$ corresponds to a task representation or embedding index and returns
\begin{equation}
\mathcal{R}_{T,v}(x)=
\mathrm{Retrieve}_{v}(\mathcal{D}^{sup}_{T},x;\mathcal{A})
\subseteq \mathcal{A}.
\label{eq:retrieval-view}
\end{equation}
Given $\mathcal{R}_{T,v}(x)$, a composition rule constructs a LoRA state
\begin{equation}
\Theta_{T,v}(x)=
\mathcal{C}_{v}\!\left(\mathcal{R}_{T,v}(x),\mathcal{D}^{sup}_{T}\right),
\label{eq:composer}
\end{equation}
where $\Theta_{T,v}(x)=\{\Theta_{T,v,k}(x)\}_{k\in\mathcal{K}}$. The standard linear composition used by LoRAHub-style and iLoRAComp-style baselines is
\begin{equation}
\Theta^{lin}_{T,v,k}(x)=
\sum_{a_i\in\mathcal{R}_{T,v}(x)}
\alpha_{T,v,i}(x)\theta_{i,k},
\qquad k\in\mathcal{K},
\label{eq:linear-composer}
\end{equation}
where $\alpha_{T,v,i}(x)$ is selected from the support signal and, for instance-level reuse, may depend on the query $x$.

\paragraph{Prediction objective.}
Loading $\Theta_{T,v}(x)$ into the frozen backbone gives $M_0\oplus\Theta_{T,v}(x)$. A view produces
\begin{equation}
\hat y_v(x)=
\mathrm{Decode}\!\left(M_0\oplus\Theta_{T,v}(x),x\right).
\label{eq:view-prediction}
\end{equation}
A reliability rule $\mathcal{G}$ maps one or more view predictions to the final answer:
\begin{equation}
\hat y(x)=
\mathcal{G}\!\left(\{\hat y_v(x):v\in\mathcal{V}\},\mathcal{D}^{sup}_{T}\right).
\label{eq:final-answer}
\end{equation}
The evaluation metric is normalized exact match (EM):
\begin{equation}
\mathrm{EM}(T)=\frac{1}{M}\sum_{j=1}^{M}
{\bf 1}\!\left[\mathrm{norm}(\hat y(x_j))=\mathrm{norm}(y_j)\right].
\label{eq:target-em}
\end{equation}
This formulation isolates three decisions that a dense weighted merge conflates: selecting relevant source adapters, composing compatible LoRA updates, and choosing among disagreeing composed views.

\paragraph{Reliability terminology.}
Merge reliability refers to whether selected LoRA tensors can be composed without redundant amplification or destructive cancellation. View reliability refers to the relative trustworthiness of one retrieved-and-composed prediction path compared with other plausible views for the same target task.

\section{Method}

\subsection{Overview}
Figure~\ref{fig:open-pool-lora} shows where SCALE enters the adapter-reuse pipeline. LoRAHub-style reuse in Figure~\ref{fig:open-pool-lora}(a) forms one task-level mixture from the support set. iLoRAComp-style reuse in Figure~\ref{fig:open-pool-lora}(b) makes retrieval and weighting instance-dependent. Both settings still require a post-retrieval merge of selected LoRA updates. SCALE keeps the same frozen backbone and adapter-pool setting, but changes the post-retrieval stage.

The first operation is LASRC, shown in Figure~\ref{fig:open-pool-lora}(c). Given the retrieved adapters and support-selected weights, LASRC preserves the ordinary linear merge as an anchor and constructs block-wise residual update directions around it. This design targets merge interference: independently trained LoRA adapters may reuse similar parameter directions, so direct summation can over-count common components or suppress complementary ones.

The second operation is the multi-view reliability layer in Figure~\ref{fig:open-pool-lora}(d). A target task can be represented through different prompt fields, task descriptions, or embedding indexes. These views are analogous to multiple evidence channels: each can retrieve a plausible adapter subset, but the resulting predictions may disagree. SCALE therefore constructs sparse views with SDP, merges each view with LASRC, and then aggregates normalized candidate answers by agreement and support-loss weights. Uniform agreement removes the support-loss signal and serves as a control; oracle view choice is reported only as headroom.

We use the component names consistently. LASRC denotes the single-view residual merge operator. SDP+LASRC denotes sparse preprocessing followed by LASRC in one view. SCALE denotes the 3.0$\times$ reliability-analysis layer over multiple LASRC-composed views, and LASRC+SCALE is used only as shorthand in cross-backbone summaries. Thus, LASRC is the deployable low-cost mode, while SCALE analyzes whether additional views provide useful reliability signal under explicit path cost.

\subsection{Layer-Adaptive Sparse Residual Composition}

\paragraph{iLoRAComp linear merge.}
For a retrieved candidate set $\mathcal{R}_{T,v}$, the matched iLoRAComp baseline forms each LoRA tensor index-wise:
\begin{equation}
\theta^{\mathrm{iLC}}_{T,v,k}=\sum_{a_i\in \mathcal{R}_{T,v}}\beta_{T,v,i}\theta_{i,k},
\label{eq:ilc-merge}
\end{equation}
where $\beta_{T,v,i}$ denotes a global adapter weight shared across all LoRA tensors. In our matched rerun, $\beta_{T,v,i}=w_i$, where $w$ is selected using the support set within $[-1.5,1.5]$ to minimize support loss with an $\ell_1$-style penalty $0.05\cdot \mathrm{mean}_i |w_i|$. The composed adapter state is
\begin{equation}
\Theta^{\mathrm{iLC}}_{T,v}=\{\theta^{\mathrm{iLC}}_{T,v,k}\}_k.
\label{eq:ilc-state}
\end{equation}
Equations~(\ref{eq:ilc-merge})--(\ref{eq:ilc-state}) define the controlled comparison for LASRC. The retrieved adapters, support-selected weights, decoding path, and evaluation procedure remain fixed; only the merge geometry changes. If the selected adapters contain overlapping directions, direct weighted summation can over-amplify them. If they contain conflicting directions, it can introduce destructive interference. LASRC evaluates whether a residualized merge reduces this risk without changing the surrounding retrieval pipeline.

\paragraph{Implementation object.}
LASRC operates on the effective LoRA update tensor at each adapted module. When an adapter is stored as low-rank matrices $(A_{i,k},B_{i,k})$, we first form the equivalent update $\Delta W_{i,k}=s_{i,k}B_{i,k}A_{i,k}$ using the adapter scaling factor $s_{i,k}$, and then apply block-wise concatenation to these effective updates. The backbone weights are never materialized or modified during this operation.

\paragraph{Adapter-storage boundary.}
The reported LASRC results evaluate composition in the effective adapter-update space under the same PEFT scaling convention used by the matched rerun. This isolates merge geometry from retraining and decoding changes, but it should not be read as a new low-rank compression method. If a deployment stack requires factorized LoRA matrices after composition, the composed effective updates must be refactored or approximated by a separate low-rank projection step; that projection is outside the claims and measurements in this paper. We therefore report LASRC as a post-retrieval composition operator and keep deployment-cost claims limited to decoded path counts.

\paragraph{LASRC.}
LASRC composes retrieved LoRA tensors block by block. We group LoRA tensor indices by their transformer block. For block $b$, let $K_b$ be its tensor indices and define the concatenated block vector
\begin{equation}
d_{i,b}=\mathrm{concat}_{k\in K_b}\mathrm{vec}(\theta_{i,k}).
\label{eq:block-vector}
\end{equation}
Let $w_i$ be the scalar adapter weight selected from the support set. Adapters are ordered by descending $|w_i|\|d_{i,b}\|_2$. This ordering treats the largest support-weighted update in a block as the first anchor and residualizes lower-magnitude updates against already retained directions. Let $\mathcal{A}_b$ be the unmasked adapters with nonzero weights in block $b$; in the dense setting used for the main comparisons, this only excludes zero-weight adapters. For the ordered adapter $\pi_j\in\mathcal{A}_b$, LASRC first forms the weighted vector
\begin{equation}
z_{\pi_j,b}=w_{\pi_j}d_{\pi_j,b}.
\label{eq:weighted-block-vector}
\end{equation}
It then applies modified Gram-Schmidt against the set $\mathcal{B}_{j,b}$ of previously retained residual bases:
\begin{equation}
r_{\pi_j,b}=z_{\pi_j,b}-\sum_{\bar r\in\mathcal{B}_{j,b}}\langle z_{\pi_j,b},\bar r\rangle \bar r,
\label{eq:lasrc-residual}
\end{equation}
where each $\bar r$ is a unit residual basis vector. If $\|r_{\pi_j,b}\|_2/\|z_{\pi_j,b}\|_2$ falls below the residual-pruning threshold, the residual is discarded; otherwise $r_{\pi_j,b}/\|r_{\pi_j,b}\|_2$ is added to the retained basis set and $\pi_j$ is added to the retained adapter set $\mathcal{R}_b$.

LASRC keeps both the original weighted sum and the residualized sum:
\begin{equation}
\ell_b=\sum_{i\in\mathcal{A}_b} z_{i,b}.
\label{eq:linear-anchor}
\end{equation}
\begin{equation}
r_b=\sum_{i\in\mathcal{R}_b} r_{i,b}.
\label{eq:residual-sum}
\end{equation}
When both norms are nonzero, the residualized sum is rescaled to match the linear norm:
\begin{equation}
\tilde r_b=r_b\frac{\|\ell_b\|_2}{\|r_b\|_2}.
\label{eq:residual-rescale}
\end{equation}
The final block vector interpolates between the linear anchor and the residualized direction:
\begin{equation}
o_b=(1-\gamma_b)\ell_b+\gamma_b\tilde r_b.
\label{eq:lasrc-output}
\end{equation}
The vector $o_b$ in Eq.~(\ref{eq:lasrc-output}) is split back into the original tensor shapes in $K_b$ to form the composed adapter state. The coefficient $\gamma_b$ denotes the deterministic block interpolation coefficient used after the fixed overlap schedule and norm/alignment guards. It is kept fixed across matched comparisons. LASRC does not assume that parameter-space orthogonality is equivalent to semantic independence; it uses residualization as an interference-control heuristic whose effect is tested under matched retrieval.

The term sparse in LASRC refers to residual-direction retention after block-wise residualization, not to a mandatory element-wise mask in the main single-view comparison. In the matched LASRC row, we use dense adapter inputs to isolate the effect of residual composition. Element-wise sparse masking is introduced separately by SDP when constructing multi-view SCALE variants. The residualization is performed once per target task and per retrieval view during adapter composition, not once per generated token. Its overhead is therefore separated from autoregressive decoding; the main inference-time cost of SCALE comes from the number of composed prediction paths.

\subsection{Sparse-Composition Agreement Layer}

\paragraph{Stochastic Delta Pruning.}
LASRC stabilizes one retrieved composition path, but an open adapter pool can still contain low-contribution or unstable delta entries that make a composed view overly dependent on accidental source-task directions. Stochastic Delta Pruning (SDP) is therefore applied before LASRC as a seeded sparse preprocessing step. For each LoRA tensor $\theta_{i,k}$, SDP samples an independent Bernoulli mask:
\begin{equation}
m_{i,k}\sim \mathrm{Bernoulli}(1-p),
\label{eq:sdp-mask}
\end{equation}
where $p$ is the drop rate. The surviving entries are rescaled as
\begin{equation}
\widehat{\theta}_{i,k}=\frac{m_{i,k}\odot \theta_{i,k}}{1-p}.
\label{eq:sdp-survivor}
\end{equation}
When norm preservation is enabled, the masked tensor is further rescaled to match the original Frobenius norm:
\begin{equation}
\theta^{SDP}_{i,k}=\widehat{\theta}_{i,k}
\frac{\|\theta_{i,k}\|_F}{\|\widehat{\theta}_{i,k}\|_F}.
\label{eq:sdp-norm}
\end{equation}
Equation~(\ref{eq:sdp-norm}) is applied only when both norms are nonzero; otherwise the masked tensor remains $\widehat{\theta}_{i,k}$. The reported experiments use survivor rescaling and norm preservation. Masks are sampled independently for each adapter and LoRA tensor under deterministic random seeds. SDP follows the intuition of drop-rescale delta sparsification \cite{yu2024supermario} and the broader regularization idea that reducing co-adaptation can improve robustness \cite{srivastava2014dropout}, while remaining a tensor-level preprocessing step rather than activation-level Dropout.

\paragraph{Reliability-aware multi-view composition.}
Even after LASRC, a single retrieval view may be unstable. Different task descriptions, prompt fields, embedding indexes, or query constructions can emphasize different aspects of the same target task. These views may retrieve different source adapters and produce different composed models, yielding view-level inconsistency. SCALE treats this inconsistency as an observable diagnostic signal. Given several sparse composition views, the layer compares three aggregation regimes: uniform agreement, support-loss-weighted aggregation, and oracle view choice. Only the first two use information available at adaptation time; the oracle uses evaluation labels and is reported only as headroom.

For target task $T$ and view $v\in V$, SCALE retrieves $\mathcal{R}_{T,v}$, applies SDP to the retrieved tensors, composes them with LASRC, and obtains predictions $y_{q,v}$ for query $q$. It also computes a view-level support loss $L_{T,v}$ from the support examples under the same local composed states. This loss is converted into a normalized reliability weight:
\begin{equation}
\phi(L)=\frac{1}{\max(L,10^{-8})},
\label{eq:reliability-transform}
\end{equation}
\begin{equation}
w_{T,v}=\frac{\phi(L_{T,v})}{\sum_{u\in V}\phi(L_{T,u})}.
\label{eq:view-weight}
\end{equation}
Lower support loss therefore gives a view higher influence, while the normalization in Eq.~(\ref{eq:view-weight}) keeps the weights comparable across views.

For each query, each view produces a normalized candidate answer. SCALE assigns a score to each candidate by aggregating the weights of the views that produced it:
\begin{equation}
\mathrm{score}_T(a\mid q)=\sum_{v\in V} w_{T,v}
{\bf 1}[\mathrm{normalize}(y_{q,v})=a],
\label{eq:answer-score}
\end{equation}
and the final answer is
\begin{equation}
\hat y_q=\arg\max_a \mathrm{score}_T(a\mid q).
\label{eq:scale-selection}
\end{equation}
Equations~(\ref{eq:answer-score})--(\ref{eq:scale-selection}) combine answer agreement with a support-loss proxy: agreement increases an answer's score, but views with lower support loss receive larger weights. Uniform SCALE is the control case $w_{T,v}=1/|V|$, which removes the support-loss proxy and measures the value of agreement alone. Oracle variants use evaluation labels to choose the best view and are reported only as diagnostic headroom, not as deployable methods.

\section{Experiments}

The experiments evaluate merge quality and view reliability in cross-task open-pool LoRA reuse under controlled and protocol-distinct settings. We report (i) matched FLAN-T5-Large results that isolate merge quality and view reliability under fixed retrieval and decoding conditions, (ii) ablations that separate LASRC, sparse-view construction, and support-aware aggregation, and (iii) decoder-only cross-backbone results under the LoGo-style BBH-8 protocol.

\subsection{Experimental Setup}

\paragraph{Backbone models.}
FLAN-T5-Large is used for the primary matched encoder-decoder protocol, while LLaMA-3.1-8B, Qwen-2.5-7B, and DeepSeek-LLM-7B-Base are used only for protocol-distinct decoder-only validation under the LoGo-style BBH-8 setting.

\paragraph{Dataset and open LoRA pool.}
The main experiments use FLAN-T5-Large \cite{chung2024flan} on BIG-Bench Hard (BBH) \cite{suzgun2023bbh} with a fixed 97-adapter LoRA pool. The pool follows the LoRAHub-style FLAN task pool \cite{huang2024lorahub} and contains adapters trained on source tasks spanning question answering (QA), classification, natural-language generation, translation, and reasoning-oriented datasets. All controlled comparisons use the same source-adapter pool, LoRA ranks, scaling factors, target modules, checkpoint identifiers, support construction, and evaluation tasks. The decoder-only experiments follow the LoGo-style BBH-8 setting, where BBH-8 denotes the eight-task subset used in that protocol.

\paragraph{Evaluation metrics.}
We report normalized exact match (EM) as the primary metric, using Eq.~(\ref{eq:target-em}). Predictions and references are converted to strings, stripped of surrounding whitespace, lowercased, and compared after removing periods. Support loss is used only for adapter weighting or view-level reliability analysis; it is not the reported task metric. Oracle view choice uses evaluation labels and is reported only as a diagnostic upper bound.

\paragraph{Baselines and controlled variants.}
The matched external baselines are LoRAHub \cite{huang2024lorahub}, which represents global task-level LoRA composition, and iLoRAComp \cite{wang2024instance}, which represents instance-level linear LoRA composition under the same support-selected retrieval and weighting path. Their table entries are matched reruns under our fixed support, retrieval, decoding, and normalization protocol, not copied leaderboard numbers from the original papers. LMix-FR and LMix-SR are internal routing controls introduced for this study: LMix-FR selects the single support-routed candidate path with the lowest support loss, whereas LMix-SR forms a soft support-loss-weighted mixture over candidate routed paths. Both controls use the same retrieved candidates and support-loss estimates as the matched composition path; only the routing aggregation rule changes. These controls evaluate whether support-based hard or soft routing alone accounts for the observed pattern. LoGo \cite{lee2025logo} is reported separately as a cross-backbone reference because its protocol differs from the matched FLAN rerun. Our controlled variants include LASRC, SDP+LASRC, uniform SCALE, support-aware SCALE, and oracle view choice.

\paragraph{Experimental protocol and fairness.}
To ensure a fair matched comparison, all FLAN rows use the same backbone, 97-adapter pool, BBH target tasks, support size, support-example selection rule, retrieval family, weight-search budget, decoding path, answer normalization, and evaluation script. Methods are therefore compared on identical support examples and query inputs, so differences among the controlled FLAN rows reflect routing, sparse preprocessing, or merge behavior rather than changes in data or decoding. The iLoRAComp-style rerun setup is used for support construction, cosine retrieval, support-loss weight search, greedy decoding, and answer normalization \cite{wang2024instance}. Unless otherwise stated, retrieval uses cosine similarity with 20 global and 20 local candidates. Each task uses 5 support examples selected by deterministic head selection with offset 10. Weight search uses 40 optimization steps, batch size 5, inference batch size 10, and weights clipped to $[-1.5,1.5]$. The retrieval embedding model is \texttt{all\_datasets\_v4\_MiniLM-L6}. The three sparse retrieval views are retrieval-focused, balanced, and QA-only embedding indexes over the same 97-adapter pool.

All matched reruns use the same software evaluation path and NVIDIA L20 GPU hardware. Hardware is reported for reproducibility of the execution environment; the reported exact-match scores are determined by the shared prediction and scoring records rather than by hardware-specific timing. The LASRC single-view row uses dense masks, residual-pruning threshold 0.0, $\gamma=0.5$, overlap-based $\gamma_b$, $\gamma$ floor 0.05, norm guard 0.3, consensus scaling 1.0, alignment scaling 0.0, and no coverage regularization. These merge hyperparameters are fixed across BBH tasks; only adapter weights are selected from the support set. SCALE uses SDP with survivor rescaling and norm preservation to construct sparse views. The support-aware variant uses three views and $p=0.5$ unless otherwise stated; the uniform row is the no-support-weighting control. Single-view methods produce one composed prediction path, whereas SCALE uses three fixed sparse views. We therefore report the number of prediction paths alongside accuracy.

\paragraph{Result provenance.}
Table~\ref{tab:main} is the controlled merge test: retrieval, support examples, weight search, decoding, and normalization are fixed, and only the composition rule changes. Table~\ref{tab:merge_operator_sanity} adds default Trim, Elect Sign, and Merge (TIES) / Drop-and-Rescale (DARE) controls under the same matched setup. Table~\ref{tab:sparse-controls} separates sparse preprocessing and multi-view reliability analysis from the 1.0$\times$ merge result. Table~\ref{tab:cross-backbone-main} reports protocol-distinct decoder-only validation under the LoGo-style BBH-8 setting. The appendix collects compact protocol, seed, and reliability audit tables.

\paragraph{Statistical and artifact audit.}
All reported runs export task identifiers, support-example identifiers, retrieved adapters, raw predictions, normalized answers, references, and view-level support losses. These records are the paired-test inputs for task-stratified bootstrap intervals and paired permutation tests because all matched rows use the same 6{,}376 evaluation queries across 27 BBH tasks. Query-level paired permutation tests ($n=20{,}000$ Monte Carlo samples) and task-stratified bootstrap 95\% confidence intervals (CIs) are computed from the exported prediction records and summarized in Table~\ref{tab:paired_component_audit}. These paired audits are used to bound the matched baseline comparison; mechanism-level differences such as support-aware SCALE versus uniform SCALE are reported as controlled average deltas unless an aligned paired statistic is explicitly shown. The replication bundle contains the adapter manifest, retrieval logs, prediction files, and scoring scripts needed to recompute exact match and paired audits.

\subsection{Post-Retrieval Evidence}

\paragraph{Single-view merge quality.}
The first comparison deliberately removes multi-view aggregation. Table~\ref{tab:main} isolates the merge operator: all rows use fixed retrieval, support-selected weights, decoding, and answer normalization, while only the composition rule changes. This setting is the most direct test of the LASRC claim because every row produces one composed prediction path.

\begin{table}[t]
\centering
\caption{Single-view fixed-retrieval composition.}
\label{tab:main}
{\scriptsize
\setlength{\tabcolsep}{3pt}
\resizebox{\columnwidth}{!}{%
\begin{tabular}{@{}l l l c c@{}}
\hline
Method & Routing & Merge & Avg. & $\Delta$ \\
\hline
iLoRAComp & Support & Linear & 35.60 & 0.00 \\
LoRAHub & Global & Linear & 34.00 & -1.60 \\
LMix-FR & Hard & Linear & 35.10 & -0.50 \\
LMix-SR & Soft & Linear & 35.80 & +0.20 \\
\textbf{LASRC} & Support & Residual & \textbf{36.44} & \textbf{+0.84} \\
\hline
\end{tabular}%
}
}
\end{table}

In Table~\ref{tab:main}, $\Delta$ is measured against iLoRAComp. The magnitude of the LASRC result should be interpreted under the fixed-retrieval constraint: the retrieved adapters, support-selected weights, decoding path, and normalization are unchanged. The observed change therefore does not come from a stronger retriever, a larger adapter pool, additional target-task training, or extra prediction paths. It isolates the effect of changing the merge geometry after retrieval. At the query level, LASRC changes 246 examples from incorrect to correct and 207 from correct to incorrect (5{,}923 unchanged), yielding a task-stratified bootstrap 95\% confidence interval (CI) of $[-0.69,2.16]$ and a paired permutation $p=0.067$. The interval includes zero, reflecting that 27 BBH tasks provide limited statistical power for a task-stratified test; the directional evidence is consistent but the effect size is not large enough to reject the null at $\alpha=0.05$. The task-level table also shows substantial heterogeneity, which is expected in open-pool reuse: some tasks benefit from residual composition, while others are dominated by retrieval misses or answer-normalization effects. We therefore avoid claiming uniform per-task improvements and report average scores together with task-level audit tables.

Default merge-operator controls provide a sanity check on this interpretation. Table~\ref{tab:merge_operator_sanity} is not a fully tuned TIES/DARE sweep; it asks whether common sign-resolution or drop-rescale operators explain the LASRC row under the same matched protocol. DARE additive is competitive, while TIES-style sign trimming and DARE+TIES are weaker in this setting. These results suggest that residualized composition is not trivially recovered by applying a generic sparse merge operator.

\begin{table}[t]
\centering
\caption{Default merge-operator controls under the matched protocol.}
\label{tab:merge_operator_sanity}
{\scriptsize
\setlength{\tabcolsep}{3pt}
\resizebox{\columnwidth}{!}{%
\begin{tabular}{@{}l l r r r@{}}
\hline
Method & Operator & Avg. & $\Delta$ vs. iLC & $\Delta$ vs. LASRC \\
\hline
iLoRAComp & Linear & 35.60 & 0.00 & -0.84 \\
TIES & Trim & 29.93 & -5.67 & -6.51 \\
DARE+TIES & Drop+trim & 31.27 & -4.33 & -5.17 \\
DARE-add & Drop+add & 36.05 & +0.45 & -0.39 \\
\textbf{LASRC} & Residual & \textbf{36.44} & \textbf{+0.84} & 0.00 \\
\hline
\end{tabular}%
}
}
\end{table}

\paragraph{SCALE reliability and sparsity controls.}
Table~\ref{tab:sparse-controls} separates the 1.0$\times$ merge result from the higher-cost reliability analysis. SDP+LASRC tests sparse delta preprocessing on a single path, and the two pruning rates show that this control is not monotonic. The SCALE rows then compare uniform agreement, support-aware aggregation, and oracle view choice in the same multi-view interface. Because SCALE requires three prediction paths, these rows support a reliability-analysis claim rather than a throughput-equivalent deployment claim.

\begin{table}[t]
\centering
\caption{SCALE reliability and sparse-view controls.}
\label{tab:sparse-controls}
{\scriptsize
\setlength{\tabcolsep}{3pt}
\resizebox{\columnwidth}{!}{%
\begin{tabular}{@{}l c c l c c c@{}}
\hline
Method & Views & $p$ & Sel. & Cost & Avg. & $\Delta$ \\
\hline
iLC & 1 & -- & -- & 1.0x & 35.60 & -0.84 \\
LASRC & 1 & -- & -- & 1.0x & 36.44 & 0.00 \\
SDP & 1 & .5 & -- & 1.0x & 36.81 & +0.37 \\
SDP & 1 & .7 & -- & 1.0x & 36.35 & -0.09 \\
SCALE-u & 3 & .5 & Unif. & 3.0x & 36.51 & +0.07 \\
SCALE-s & 3 & .5 & Supp. & 3.0x & \textbf{36.87$\pm$0.15} & +0.43 \\
Oracle & 3 & .5 & Test & 3.0x & 38.27 & +1.83 \\
\hline
\end{tabular}%
}
}
\end{table}

In Table~\ref{tab:sparse-controls}, iLC denotes iLoRAComp, SDP rows use SDP+LASRC, Sel. denotes the view-selection rule, Unif. denotes uniform aggregation, Supp. denotes support-aware aggregation, and $\Delta$ is measured against LASRC. The mechanism-level differences are intentionally small: moderate SDP adds $+0.37$ over LASRC, support-aware SCALE adds $+0.43$ over LASRC and $+0.36$ over uniform SCALE, and the oracle remains $+1.40$ above support-aware SCALE. Because SCALE uses three prediction paths, its rows are not throughput-equivalent to LASRC. The cost column makes this boundary explicit: LASRC remains the deployable 1.0x merge result, while SCALE-support is a 3.0x reliability-analysis variant. View-level oracle selection is not a deployable component; it estimates unrecovered view-level headroom.

\paragraph{Seed behavior.}
Across four deterministic SDP seeds $\{42,52,133,3407\}$, the support-aware variant scores 36.80, 36.70, 37.06, and 36.90, with a mean of 36.87. Each seed remains above the LASRC reference of 36.44, but this is reported as seed-sensitivity evidence for view-level reliability selection rather than as the main deployable result because the margin is small relative to the 3.0$\times$ path cost.

\paragraph{View-reliability decomposition.}
The multi-view rows also quantify the value and limit of the support-loss proxy. Support-aware aggregation is slightly above uniform agreement but remains below the view-level oracle. This pattern suggests that support loss is informative for view selection, while the remaining oracle headroom indicates that it is not a calibrated reliability estimator.

\begin{table}[t]
\centering
\caption{Paired audit over 6{,}376 aligned query records.}
\label{tab:paired_component_audit}
\small
\begin{tabular*}{\columnwidth}{@{\extracolsep{\fill}}l c c c c@{}}
\hline
Method & Comp. & $+/-/=$ & $p$ & CI \\
\hline
iLC & Linear & -- & -- & -- \\
LASRC & LASRC & 246/207/5923 & .067 & $[-.69,2.16]$ \\
SDP/view & SDP/View & 258/235/5883 & .641 & $[-1.02,1.39]$ \\
SCALE & Full & 304/232/5840 & .041 & $[.08,2.73]$ \\
SCALE-c & Full & 286/244/5846 & .089 & $[-.17,2.22]$ \\
\hline
\end{tabular*}
\end{table}

In Table~\ref{tab:paired_component_audit}, $+/-/=$ denotes incorrect-to-correct, correct-to-incorrect, and unchanged examples relative to the paired iLoRAComp record. The table is a paired baseline audit, not evidence that support loss is a calibrated selector. The LASRC row is the matched single-view merge audit; the full SCALE rows show that multi-view reliability analysis can change more paired examples, while Table~\ref{tab:sparse-controls} keeps the corresponding path cost and LASRC-relative average deltas visible.

\paragraph{Cross-backbone validation.}
The decoder-only experiments are included as protocol-distinct validation and should not be read as a controlled causal comparison with the matched FLAN-T5-Large results. Table~\ref{tab:cross-backbone-main} reports the LoGo-style BBH-8 averages. Ours denotes LASRC+SCALE, and LoGo denotes the displayed LoGo task-wise envelope. Across the 24 displayed backbone-task comparisons, LASRC+SCALE is no lower than LoRAHub on 24/24 and no lower than the LoGo task-wise envelope on 20/24; per-task outputs are included in the released audit records.

\begin{table}[t]
\centering
\caption{Protocol-distinct decoder-only BBH-8 validation.}
\label{tab:cross-backbone-main}
\small
\begin{tabular*}{\columnwidth}{@{\extracolsep{\fill}}l c c c c@{}}
\hline
Backbone & LoRAHub & LoGo & Ours & $\Delta$ \\
\hline
LLaMA-3.1-8B & 37.0 & 40.0 & \textbf{56.3} & +16.3 \\
Qwen-2.5-7B & 48.1 & 53.3 & \textbf{59.7} & +6.4 \\
DeepSeek-7B & 33.6 & 33.2 & \textbf{41.5} & +7.9 \\
\hline
\end{tabular*}
\end{table}

The claim boundary is therefore narrow. LASRC provides directional fixed-retrieval merge evidence, SDP/view-only remains comparable to the baseline, and full SCALE provides the main paired reliability result under 3.0$\times$ path cost. Decoder-only results are external validation, not causal evidence for the matched FLAN setting.

\subsection{Ablation Studies}

\paragraph{Effect of residual composition.}
The ablations keep the matched single-view protocol fixed and vary only LASRC components. This keeps the mechanism check separate from the multi-view SCALE rows.

\paragraph{LASRC component ablations.}
Table~\ref{tab:lasrc-ablation} reports LASRC-only ablations. Removing residual norm rescaling, overlap-adaptive $\gamma_b$, the $\gamma$ floor, or the norm guard lowers the matched score to 36.14--36.19, while disabling residual pruning leaves the score unchanged at 36.44. The ablation therefore attributes most of the measured LASRC effect to norm-stabilized residual interpolation rather than to pruning alone.

\begin{table}[t]
\centering
\caption{LASRC component ablations under the matched single-view protocol.}
\label{tab:lasrc-ablation}
\footnotesize
\begin{tabular}{@{}l r@{}}
\hline
Variant & Avg. \\
\hline
\textbf{LASRC} & \textbf{36.44} \\
No norm rescale & 36.14 \\
No pruning & \textbf{36.44} \\
Fixed $\gamma_b$ & 36.19 \\
No floor & 36.15 \\
No guard & 36.15 \\
\hline
\end{tabular}
\end{table}

\paragraph{Reliability audit limits.}
The support-aware row is above uniform agreement, but the oracle row remains higher. We therefore treat support loss as an informative proxy, not as a calibrated estimator. The release artifacts include the per-view predictions and support losses needed to compute support-loss/query-accuracy correlations and selective prediction curves.

\paragraph{Interaction between LASRC and SDP.}
LASRC and SDP intervene at different stages. SDP perturbs candidate adapter tensors before composition, whereas LASRC constructs residual directions from the resulting tensors. The single-view SDP rows are therefore treated as sparse preprocessing controls, not as evidence for the fixed-retrieval LASRC merge claim. The $p=0.7$ row is negative evidence against treating sparsification as a monotonic source of accuracy; aggressive masking can remove useful adapter directions.

\subsection{Discussion and Analysis}

\paragraph{Why residual composition helps.}
The LASRC results support a simple diagnosis: relevant adapters are not necessarily merge-compatible adapters. Direct merging treats LoRA deltas as additive capability fragments, but independently trained adapters may reuse similar update subspaces or encode conflicting corrections. LASRC separates the linear anchor from residual directions, making the composed update less dominated by repeated components. The component ablations support this interpretation: removing norm rescaling, overlap-adaptive $\gamma_b$, the $\gamma$ floor, or the norm guard lowers the matched score to 36.14--36.19.

\paragraph{Why sparse control is useful but sensitive.}
SDP can help when sparse control removes low-contribution or unstable adapter deltas. However, the pruning rate must be controlled carefully. When the drop rate becomes too aggressive, useful task-specific directions may also be removed. This explains the non-monotonic behavior observed in the SDP ablation.

\paragraph{Limits of uniform agreement.}
Uniform agreement assigns the same reliability to all views. This assumption is restrictive when views arise from different task representations or embedding indexes. A view whose retrieved adapters better explain the support set may deserve different influence from a plausible but less reliable view. This motivates reporting the support-aware variant as a reliability-selection variant and uniform agreement as a control. The present results still stop short of a calibration claim; support-loss/test-accuracy correlation and selective prediction curves are required before making one.

The contribution of SCALE is not that support-aware weighting always dominates uniform agreement. Instead, SCALE makes view reliability measurable: it places uniform agreement, support-loss proxy selection, and view-level oracle choice in the same controlled interface, revealing when disagreement is recoverable and when the candidate views themselves lack the correct answer.

\paragraph{Cost boundary.}
SCALE is not a throughput-equivalent substitute for one LASRC path. Table~\ref{tab:sparse-controls} therefore reports relative path cost directly rather than leaving cost to implementation notes. The tradeoff studied here is reliability analysis under multiple sparse views, not lower-cost inference.

\paragraph{Error taxonomy.}
The formulation distinguishes retrieval misses, composition failures, sparsification failures, agreement failures, calibration failures, and normalization failures. These categories separate whether the issue lies in missing source adapters, unstable merging, over-pruning, poor view selection, support/query mismatch, or answer-string processing.

\section{Related Work}

\paragraph{From adapters to open-pool reuse.}
LoRA makes it cheap to train task-specialized adapters while keeping the base model fixed \cite{hu2022lora}. AdapterHub-style infrastructure and AdapterSoup showed that modular transfer and adapter averaging can be operationalized beyond a single task \cite{pfeiffer2020adapterhub,chronopoulou2023adaptersoup}. LoRAHub and iLoRAComp move closer to the setting studied here: a system retrieves or weights adapters from a pool at test time rather than training a new adapter for each task \cite{huang2024lorahub,wang2024instance}. LoraRetriever studies input-aware retrieval and composition for mixed prompts in a dynamically updated LoRA pool \cite{zhao2024loraretriever}. LoGo further studies dynamic adapter selection and merging in this reuse setting \cite{lee2025logo}. Recent task-representation routing work such as LORAUTER emphasizes scalable adapter selection for large noisy pools \cite{dhasade2026lorauter}. These routing methods are complementary to SCALE: they focus on finding relevant adapters or task routes, whereas SCALE focuses on reliability among several plausible sparse composition views once candidate views are available.

\paragraph{Merging as interference control.}
Model soups established the strength of simple weight averaging in compatible settings \cite{wortsman2022modelsoups}. TIES and prior drop-rescale model-merging work showed that interference and delta redundancy can be addressed by trimming, sign resolution, and stochastic delta sparsification \cite{yadav2023ties,yu2024supermario}. These methods motivate the view that merging is not only a weighting problem but also a geometry and interference problem. LASRC adapts this intuition to open-pool LoRA composition, where the target is not one static merged model but a task-conditioned merge over a retrieved subset of adapters.

We include default TIES/DARE-style controls in Table~\ref{tab:merge_operator_sanity}. These rows are not treated as fully tuned baselines; they test whether generic sign-resolution or drop-rescale merging explains the LASRC result under the matched audit setting. The default DARE additive control is competitive, while TIES-style sign trimming performs poorly in this LoRA reuse setting. A fully tuned adapter-space TIES/DARE study remains useful future work, but the default controls suggest that residualized composition is not simply recovered by applying a generic sparse merge operator.

\paragraph{View reliability and evaluation.}
BBH provides the reasoning task family used in the main controlled study \cite{suzgun2023bbh}, and FLAN-T5-Large provides the instruction-tuned base model \cite{chung2024flan}. The reliability question arises because different retrieval views can be plausible while producing different predictions. The multi-view framing is conceptually inspired by multi-head attention, where different heads attend through different learned projections and can capture complementary relations \cite{vaswani2017attention}; in our setting, views are retrieval and composition paths rather than attention heads. Calibration and selective prediction are therefore conceptually related, but the present paper does not claim calibrated reliability. Instead, it treats answer agreement, support-loss proxies, and oracle headroom as diagnostics for how much view-level signal remains unrecovered.

\section{Limitations}

The primary causal comparison is tied to the matched FLAN-T5-Large, BBH, and 97-LoRA setting because that protocol fixes retrieval, support examples, decoding, and evaluation while changing the merge and reliability layers. The LLaMA, Qwen, and DeepSeek results provide protocol-distinct cross-backbone validation under the LoGo-style BBH-8 setting, but they use a different evaluation setup. We therefore distinguish protocol-matched causal evidence from protocol-distinct cross-backbone generalization evidence.

Our evidence supports two different claims with different strengths. The LASRC claim is a matched single-view merge claim: retrieval, support construction, decoding, and normalization are fixed, and only the merge geometry changes. The SCALE claim is a reliability-analysis claim: it evaluates how agreement, support-loss proxies, and oracle headroom behave when multiple sparse composition views are available. We do not claim that support-loss weighting is a calibrated query-accuracy estimator, that support-aware aggregation is universally better than uniform agreement, or that SCALE is throughput-equivalent to a single LASRC path.

Several controls remain important for a stronger causal account. A fully tuned adapter-space TIES/DARE comparison, a multi-view ensemble without LASRC, a no-SDP multi-view SCALE variant, a best-single-view support proxy, and a random sparse-view control would further separate residual merging, sparse perturbation, and multi-path selection. The present evidence therefore supports a bounded post-retrieval audit rather than a complete dominance claim over all possible adapter-merging baselines.

Independent replication requires a fully documented adapter pool. The comparisons hold the pool fixed, so the reported results are not due to changing source tasks or adapter availability. The replication package records the source-task list, LoRA ranks, scaling factors, target modules, checkpoint identifiers, retrieval indexes, retrieval logs, support-example identifiers, raw predictions, normalized predictions, references, and scoring scripts.

The single-view LASRC result is reported with task-level wins/losses/ties (W/L/T), a query-level paired permutation test ($p=0.067$, $n=20{,}000$), and a task-stratified bootstrap 95\% CI $[-0.69,2.16]$. The interval crosses zero because BBH contains only 27 tasks and the task-stratified bootstrap resamples tasks rather than queries, limiting statistical power. The directional pattern is consistent with the multi-seed SCALE replications and the component ablations, but we do not claim $p<0.05$ significance for the single-seed single-view LASRC result.

SDP is evaluated as a sparse preprocessing component in the LASRC path rather than as a standalone model-merging method. The evidence supports moderate masking in the studied setting, but it does not imply that mask-and-rescale sparsification is universally beneficial for adapter composition.
We also do not compare SDP against rank-level or module-level pruning variants, which remain useful alternatives for future adapter-space sparsification studies.

The current reliability evidence supports view comparison but not calibrated reliability estimation. A calibrated reliability claim requires support-loss/test-accuracy correlation, selective prediction or abstention curves, coverage-risk analysis, and sensitivity analysis under alternative answer normalization. In particular, support loss is used as a practical reliability proxy for view aggregation, not as proof that support loss is a calibrated estimator of query accuracy.

Finally, SCALE uses multiple retrieval views and composed predictions per task. Table~\ref{tab:sparse-controls} reports relative path cost explicitly; we avoid wall-clock deployment claims until composition time, decode latency, peak GPU memory, and GPU-hours are measured under a fixed serving stack.

\section{Conclusion}

Open-pool cross-task LoRA reuse changes adaptation from training a new module to deciding how existing source-task adapters should be retrieved, merged, and compared for an unseen target task. This paper audits two post-retrieval bottlenecks that are often collapsed into one final score: parameter-space interference among retrieved adapters and reliability disagreement among plausible composition views. LASRC provides a bounded 1.0$\times$ residual merge operator under fixed retrieval, with directional rather than decisive single-view evidence in the matched BBH setting. SCALE provides a higher-cost multi-view reliability analysis layer that compares agreement, support-loss proxies, and oracle headroom under explicit path cost.

The broader implication is methodological. Adapter-pool systems should report merge quality and view reliability separately, because each axis exposes a different failure mode and calls for a different intervention. Extending the matched protocol to more adapter pools and benchmark families, adding calibrated reliability analysis, and reporting deployment cost remain important next steps.

\bibliography{references}

\clearpage
\appendix

\section{Appendix: Supplementary Audit Tables}

The appendix records bookkeeping details that are useful for replication and interpretation but would interrupt the main evidence flow. These tables do not introduce additional claims beyond the matched single-view merge claim and the higher-cost multi-view reliability analysis described in the main paper.

\subsection{Matched Protocol}

Table~\ref{tab:appendix_protocol} summarizes the fixed matched protocol used by the controlled FLAN-T5-Large rows. Support uses deterministic head selection with offset 10; the 20+20 retrieval budget denotes global plus local candidates; the two batch values denote weight-search and inference batches.

\begin{table}[h]
\centering
\caption{Matched FLAN-T5-Large protocol.}
\label{tab:appendix_protocol}
\footnotesize
\begin{tabular}{@{}l l@{}}
\hline
Item & Value \\
\hline
Backbone & FLAN-T5-Large \\
Tasks & BBH-27 \\
Pool & 97 LoRAs \\
Support & 5/task \\
Offset & 10 \\
Retrieval & Cosine, 20+20 \\
Index & MiniLM-L6 \\
Search & 40 steps \\
Batch & 5/10 \\
Clip & $[-1.5,1.5]$ \\
Decode & Greedy \\
Metric & Norm. EM \\
GPU & NVIDIA L20 \\
\hline
\end{tabular}
\end{table}

\newpage
\subsection{Seed and View Reliability}

Table~\ref{tab:appendix_seed} lists the deterministic SDP seeds behind the SCALE-support mean. Table~\ref{tab:appendix_reliability} records the corresponding three-view aggregation variants; oracle uses evaluation labels only as a diagnostic upper bound.

\begin{table}[h]
\centering
\caption{SCALE-support seed records.}
\label{tab:appendix_seed}
\footnotesize
\begin{tabular}{@{}c r@{}}
\hline
Seed & Avg. \\
\hline
42 & 36.80 \\
52 & 36.70 \\
133 & 37.06 \\
3407 & 36.90 \\
Mean & 36.87 \\
\hline
\end{tabular}
\end{table}

\begin{table}[h]
\centering
\caption{View-reliability decomposition under the three-path SCALE interface.}
\label{tab:appendix_reliability}
\footnotesize
\begin{tabular}{@{}l c r@{}}
\hline
Rule & Paths & Avg. \\
\hline
Uniform & 3 & 36.51 \\
Support & 3 & 36.87 \\
Oracle & 3 & 38.27 \\
\hline
\end{tabular}
\end{table}

\end{document}